\documentclass{article}
\usepackage{amsfonts}
\usepackage{bm}
\usepackage{multirow}
\usepackage[table]{xcolor}
\usepackage{xspace}
\usepackage{pifont}
\newcommand{\cmark}{\ding{51}}
\usepackage{xcolor}
\usepackage{spconf,amsmath,graphicx}
\usepackage[hidelinks]{hyperref}

\newcommand{\method}{UPL\xspace}

\title{Uncertainty-aware Prototype Learning with Variational Inference for Few-shot Point Cloud Segmentation}

\name{Yifei Zhao$^{1}$\thanks{$^{1}$ and $^{2}$ contributed equally to this work; $^{3}$ is the corresponding author.}\qquad
      Fanyu Zhao$^{2}$\footnotemark[1]\qquad
      Yinsheng Li$^{3}$\footnotemark[1]}
\address{Fudan University, Shanghai, China\\
\texttt{\{yfzhao19, fyzhao20, liys\}@fudan.edu.cn}}

\begin{document}
\ninept
\maketitle

\begin{abstract}
Few-shot 3D semantic segmentation aims to generate accurate semantic masks for query point clouds with only a few annotated support examples. Existing prototype-based methods typically construct compact and deterministic prototypes from the support set to guide query segmentation. However, such rigid representations are unable to capture the intrinsic uncertainty introduced by scarce supervision, which often results in degraded robustness and limited generalization.
In this work, we propose \textbf{UPL} (Uncertainty-aware Prototype Learning), a probabilistic approach designed to incorporate uncertainty modeling into prototype learning for few-shot 3D segmentation. Our framework introduces two key components. First, UPL introduces a dual-stream prototype refinement module that enriches prototype representations by jointly leveraging limited information from both support and query samples. Second, we formulate prototype learning as a variational inference problem, regarding class prototypes as latent variables. This probabilistic formulation enables explicit uncertainty modeling, providing robust and interpretable mask predictions.
Extensive experiments on the widely used ScanNet and S3DIS benchmarks show that our UPL achieves consistent state-of-the-art performance under different settings while providing reliable uncertainty estimation.
The code is available at \href{https://fdueblab-upl.github.io/}{\textcolor{blue}{\textbf{fdueblab-upl.github.io}}}.

\end{abstract}


\begin{figure*}[t]
  \centering
  \includegraphics[width=0.85\textwidth]{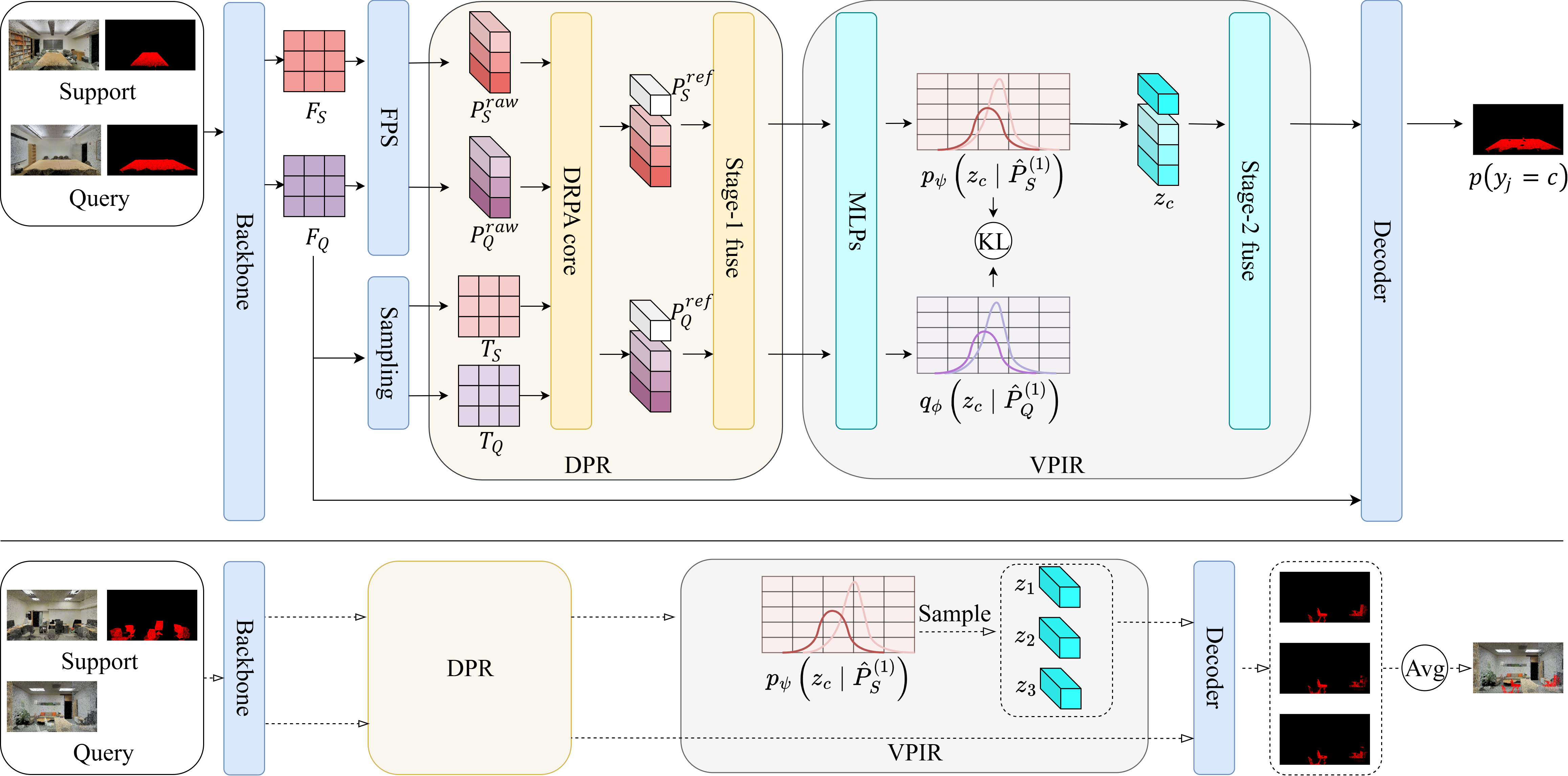}
  \vspace{-0.3cm} 
  \caption{Overview of our \textbf{UPL} framework. UPL consists of two primary modules: (i) the Dual-stream Prototype Refinement (DPR) module, which enhances prototype discriminability by leveraging mutual information between support and query sets, and (ii) the Variational Prototype Inference Regularization(VPIR) module, which models class prototypes as latent variables to capture uncertainty and enable probabilistic inference. The solid lines indicate the training process, while the dashed lines represent the inference process.}
  \vspace{-0.5cm} 
  \label{fig:UPL_framework}
\end{figure*}

\section{Introduction}

3D point cloud segmentation aims to assign a semantic label to each point and has broad applications in autonomous driving~\cite{technologies13080322, bi2025multi}, robotics~\cite{DBLP:journals/corr/abs-2009-08920, guo2020deep}, and AR/VR~\cite{pierdicca2020point,zhang2023deep}.
Recent advances, such as Mask3D~\cite{schult2023mask3d} and OneFormer3D~\cite{kolodiazhnyi2024oneformer3d}, have achieved remarkable results when trained on large-scale annotated datasets.
However, constructing such datasets with dense point-level annotations is both costly and time-consuming~\cite{wang2024survey,sun2024review}.
To alleviate this issue, few-shot point cloud semantic segmentation (FS-PCS) has been introduced, aiming to generalize segmentation models to novel classes with only a limited number of labeled samples~\cite{xu2023generalized,xu2023towards}.

Generally, FS-PCS follows an episodic meta-learning paradigm, where each episode consists of a support set with a few annotated scenes and a query set. 
Most existing FS-PCS methods adopt a prototype-based approach, in which prototypes derived from the support scenes are subsequently used to guide the segmentation of the query scenes~\cite{zhao2021few,zhang2023few,an2024rethinking}.
Building on this paradigm, recent works incorporate advanced techniques, such as graph convolutions~\cite{jiang2019semi} and attention mechanisms~\cite{vaswani2017attention}, to improve segmentation performance.
Despite their success, these methods construct deterministic prototypes to perform segmentation, neglecting the inherent uncertainty in few-shot scenarios with scarce supervision.
To this end, such deterministic prototypes lead to two key drawbacks:
(1) deterministic prototypes often fail to produce reliable segmentation when facing large intra-class variability and strong inter-class similarity~\cite{he2023prototype}. 
Addressing this challenge requires the model to generate more robust and discriminative prototypes that can generalize effectively under limited supervision.
(2) deterministic prototypes are incapable of modeling the uncertainty of predictions.
Uncertainty estimation improves model interpretability and reliability, especially with few samples.
While probabilistic modeling has been preliminarily explored for few-shot 2D segmentation~\cite{wang2021variational} and uncertainty estimation studied in point cloud segmentation~\cite{liu2025probabilisticinteractive3dsegmentation}, FS-PCS methods have largely neglected uncertainty modeling.

To address these challenges, we propose UPL, a probabilistic framework for few-shot 3D segmentation that enables uncertainty-aware prototype learning.
Specifically, UPL introduces two main components:
(i) a dual-stream prototype refinement (DPR) module, which integrates both support and query information through channel-wise attention. To this end, this module produces richer and more discriminative prototypes, which help mitigate intra-class variability and inter-class similarity.
(ii) a variational prototype inference regularization (VPIR) module. It formulates prototype learning as a variational inference problem, where class prototypes are treated as latent variables. During model training, the Kullback-Leibler divergence aligns the prior prototype distribution derived from support data with the posterior prototype distribution derived from query data.
During inference, by sampling multiple Monte Carlo samples from prior prototype distribution, our UPL yields not only robust mask predictions but also reliable uncertainty estimation.

To the best of our knowledge, this is the first work that models uncertainty in few-shot point cloud segmentation.
We conduct extensive experiments on two widely used benchmarks, S3DIS\cite{armeni20163d} and ScanNet\cite{dai2017scannet}, under different few-shot settings. The results demonstrate that UPL consistently achieves state-of-the-art performance while providing robust and interpretable uncertainty estimation.
Our main contributions are summarized as follows:

\begin{itemize}
\item We propose UPL, a probabilistic framework for FS-PCS that formulates few-shot learning as variational inference, modeling class prototypes as latent variables to enable uncertainty-aware 3D segmentation.

\item We introduce a dual-stream prototype refinement module, which effectively integrates information from both support and query sets, producing more discriminative prototypes.

\item Extensive experiments on S3DIS and ScanNet show that UPL achieves consistent state-of-the-art performance and provides reliable uncertainty estimation for FS-PCS tasks.
\end{itemize}

\section{Preliminaries}
We follow the episodic $N$-way $K$-shot setting, with disjoint base $\mathcal{C}_{\mathrm{base}}$ and novel $\mathcal{C}_{\mathrm{novel}}$ classes~\cite{vinyals2016matching, snell2017prototypical}. Each episode includes a support set $\{(\mathbf{Pt}^s_i,\mathbf{Y}^s_i)\}_{i=1}^{NK}$ and a query set $\{(\mathbf{Pt}^q,\mathbf{Y}^q)\}$. Each point cloud $\mathbf{Pt} \in \mathbb{R}^{n\times d_{pt}}$ is encoded via a backbone $f_\theta$ into per-point features:
\begin{equation}
\mathbf{F}=f_\theta(\mathbf{Pt}) \in \mathbb{R}^{n \times d_f}.
\end{equation}
Class prototypes are constructed by applying Farthest Point Sampling (FPS)~\cite{zhao2021few} to the feature set $\mathbf{F}$. For each class $c$, FPS selects a set of representative points, and their features are averaged to obtain sub-prototypes:
\begin{equation}
\mathbf{p}_c^{raw} = \frac{1}{|\mathbf{Pt}_c|} \sum_{i \in \mathbf{Pt}_c} \mathbf{F}_i,
\end{equation}
where $\mathbf{Pt}_c$ denotes the set of points assigned to class $c$ by FPS. The set of all class prototypes is denoted as $\mathbf{P}^{raw} = [\mathbf{p}_1^{raw}, \ldots, \mathbf{p}_C^{raw}] \in \mathbb{R}^{C \times d_p}$. If the number of available points is less than the desired sample size, zero-padding is used.

\textbf{Notation:} $\mathbf{P}_*^{raw}$ and $\mathbf{P}_*^{ref}$denote the collections of raw, refined prototypes for all classes in the support set, where $*$ denotes either $S$ or $Q$. $\mathbf{Z}$ denotes variational prototypes.

\begin{table*}[t]
  \centering
  \small
  \caption{Few-shot 3D semantic segmentation mIoU (\%) on S3DIS\cite{armeni20163d} and ScanNet\cite{dai2017scannet} under $N$-way $K$-shot settings, following the standardized CoSeg benchmark~\cite{an2024rethinking}. S0 / S1: two split seeds; Mean: average. Best in bold.}
  \label{tab:main_results}
  \resizebox{\linewidth}{!}{
  \begin{tabular}{l|c|ccc|ccc|ccc|ccc}
    \hline
    \multicolumn{1}{c|}{{\multirow{2}{*}{}}} & \textbf{\multirow{2}{*}{Methods}} & \multicolumn{3}{c|}{1-way 1-shot} & \multicolumn{3}{c|}{1-way 5-shot} & \multicolumn{3}{c|}{2-way 1-shot} & \multicolumn{3}{c}{2-way 5-shot} \\
    & & \textbf{S0} & \textbf{S1} & \textbf{Mean} & \textbf{S0} & \textbf{S1} & \textbf{Mean} & \textbf{S0} & \textbf{S1} & \textbf{Mean} & \textbf{S0} & \textbf{S1} & \textbf{Mean} \\
    \hline
    \multirow{5}{*}[0pt]{\cellcolor{white}\textbf{S3DIS}\cite{armeni20163d}} & \textbf{AttMPTI}\cite{zhao2021few} & 36.32 & 38.36 & 37.34 & 46.71 & 42.70 & 44.71 & 31.09 & 29.62 & 30.36 & 39.53 & 32.62 & 36.08 \\
    & \textbf{QGE}\cite{ning2023boosting} & 41.69 & 39.09 & 40.39 & 50.59 & 46.41 & 48.50 & 33.45 & 30.95 & 32.20 & 40.53 & 36.13 & 38.33 \\
    & \textbf{QPGA}\cite{he2023prototype} & 35.50 & 35.83 & 35.67 & 38.07 & 39.70 & 38.89 & 25.52 & 26.26 & 25.89 & 30.22 & 32.41 & 31.32 \\
    & \textbf{Coseg}\cite{an2024rethinking} & 46.31 & 48.10 & 47.21 & 51.40 & \textbf{48.68} & 50.04 & 37.44 & 36.45 & 36.95 & \textbf{42.27} & 38.45 & 40.36 \\
    & \cellcolor{gray!20}\textbf{UPL (ours)} & \cellcolor{gray!20}\textbf{48.18} & \cellcolor{gray!20}\textbf{49.02} & \cellcolor{gray!20}\textbf{48.60} & \cellcolor{gray!20}\textbf{55.92} & \cellcolor{gray!20}48.53 & \cellcolor{gray!20}\textbf{52.22} & \cellcolor{gray!20}\textbf{38.13} & \cellcolor{gray!20}\textbf{37.44} & \cellcolor{gray!20}\textbf{37.79} & \cellcolor{gray!20}41.78 & \cellcolor{gray!20}\textbf{41.96} & \cellcolor{gray!20}\textbf{41.87} \\
    \hline
    \multirow{5}{*}{\textbf{ScanNet}\cite{dai2017scannet}} & \textbf{AttMPTI}\cite{zhao2021few} & 34.03 & 30.97 & 32.50 & 39.09 & 37.15 & 38.12 & 25.99 & 23.88 & 24.94 & 30.41 & 27.35 & 28.88 \\
    & \textbf{QGE}\cite{ning2023boosting} & 37.38 & 33.02 & 35.20 & 45.08 & 41.89 & 43.49 & 26.85 & 25.17 & 26.01 & 28.35 & 31.49 & 29.92 \\
    & \textbf{QPGA}\cite{he2023prototype} & 34.57 & 33.37 & 33.97 & 41.22 & 38.65 & 39.94 & 21.86 & 21.47 & 21.67 & 30.67 & 27.69 & 29.18 \\
    & \textbf{Coseg}\cite{an2024rethinking} & 41.73 & 41.82 & 41.78 & 48.31 & 44.11 & 46.21 & 28.72 & 28.83 & 28.78 & 35.97 & 33.39 & 34.68 \\
    & \cellcolor{gray!20}\textbf{UPL (ours)} & \cellcolor{gray!20}\textbf{43.13} & \cellcolor{gray!20}\textbf{42.87} & \cellcolor{gray!20}\textbf{43.00} & \cellcolor{gray!20}\textbf{48.48} & \cellcolor{gray!20}\textbf{45.18} & \cellcolor{gray!20}\textbf{46.83} & \cellcolor{gray!20}\textbf{32.09} & \cellcolor{gray!20}\textbf{32.68} & \cellcolor{gray!20}\textbf{32.39} & \cellcolor{gray!20}\textbf{39.65} & \cellcolor{gray!20}\textbf{37.15} & \cellcolor{gray!20}\textbf{38.40} \\
    \hline
  \end{tabular}
  }
  \vspace{-0.5cm} 
\end{table*}

\section{Methodology}

\subsection{Overview}
As illustrated in Fig.~\ref{fig:UPL_framework}, the proposed framework consists of two key components: (i) a Dual-stream Prototype Refinement (DPR) module, which refines raw prototypes derived from both support and query sets using shared token attention to enhance their discriminative power; and (ii) a Variational Prototype Inference Regularization(VPIR) module, which models class prototypes as latent variables to capture uncertainty and enable probabilistic inference. Specifically, the DPR module generates refined prototypes by leveraging mutual information between support and query features, reducing intra-class variability and inter-class confusion. The VPIR module formulates prototype learning as a variational inference problem, where class prototypes are treated as Gaussian latent variables. During training, the posterior distribution aligns with the prior via KL divergence, while during inference, multiple samples from the prior distribution are used to generate robust predictions. Finally, the framework integrates raw, refined, and variational prototypes through a two-stage fusion process to produce the final representation, enabling accurate and uncertainty-aware predictions.

\subsection{Dual-stream Prototype Refinement}

We introduce the \textit{Dual-stream Prototype Refinement} (DPR) module to refine support and query prototypes via mutual interaction. Let $\mathbf{T}_S, \mathbf{T}_Q \in \mathbb{R}^{B \times d \times N}$ denote the support and query token sets, and $\mathbf{P}_S^{raw}, \mathbf{P}_Q^{raw}$ be the corresponding raw class prototypes, obtained via masked average pooling over labeled points. DPR performs two symmetric value-substitution passes using shared attention to refine both streams concurrently:
\begin{equation}
\label{eq:DPR}
\mathbf{P}_S^{ref}, \mathbf{P}_Q^{ref} = \mathrm{DPR}(\mathbf{T}_Q, \mathbf{T}_S, \mathbf{P}_S^{raw}, \mathbf{P}_Q^{raw}),
\end{equation}
where attention weights are computed in a channel-wise manner via scaled dot-product between tokens.

These weights modulate the prototype update through a value-substitution operation, followed by a 1D convolution and residual connection:
\begin{equation}
\label{eq:refine}
\mathbf{P}_*^{ref} = \mathrm{LN}\left( \mathbf{P}_*^{raw} + \mathrm{Conv1d}(\mathbf{A} \cdot \mathrm{Linear}(\mathbf{P}_*^{raw})) \right),
\end{equation}
where $* \in \{S, Q\}$. Layer normalization and learned projections enhance optimization stability and feature expressiveness.

To aggregate refined features with raw ones, we apply an adaptive fusion gate:
\begin{equation}
\label{eq:fuse}
\hat{\mathbf{p}}_c^{(1)} = (1 - \alpha_c) \mathbf{p}_c^{raw} + \alpha_c \mathbf{p}_c^{ref}, \quad \alpha_c = \sigma([\mathbf{p}_c^{raw} \Vert \mathbf{p}_c^{ref}]),
\end{equation}
where $\sigma(\cdot)$ denotes the sigmoid activation. This allows the model to control class-wise refinement adaptively.




\subsection{Variational Prototype Inference Regularization}

To model uncertainty and intra-class variation, we treat each class prototype as a latent variable drawn from a Gaussian distribution. Specifically, we define a prior $p_\psi(z_c)$ based on fused support features $\hat{\mathbf{p}}_c^{(1)}$, and a posterior $q_\phi(z_c)$ based on query features $\mathbf{p}_c^{q}$. Both distributions are parameterized by MLPs that produce the corresponding mean and variance.

During training, we use the reparameterization trick to sample prototypes from the prior, i.e., $z_c = \mu_c^{pr} + \sigma_c^{pr} \odot \epsilon$, where $\epsilon \sim \mathcal{N}(\mathbf{0}, \mathbf{I})$. A KL divergence is computed between the posterior and prior:
\begin{equation}
\mathcal{L}_{\mathrm{KL}} = \sum_c \mathrm{KL}(q_\phi(z_c) \,\|\, p_\psi(z_c)),
\label{eq:kl}
\end{equation}
which regularizes the distribution during optimization.

To integrate the sampled prototype into the final representation, we apply a learnable fusion gate between the deterministic prototype $\hat{\mathbf{p}}_c^{(1)}$ and the sampled $z_c$:
\begin{equation}
\hat{\mathbf{p}}_c^{(2)} = (1 - \beta_c)\hat{\mathbf{p}}_c^{(1)} + \beta_c z_c,
\label{eq:varfusion}
\end{equation}
where $\beta_c$ is obtained via a sigmoid gate over the concatenated inputs.

At test time, multiple samples $\{z_c^{(t)}\}_{t=1}^T$ are drawn from the prior to form an ensemble prediction, and we average the logits over $T$ stochastic forward passes.

\subsection{Uncertainty Estimation and Learning Objective}

To quantify predictive uncertainty, we adopt a Monte Carlo ensemble approach using $T$ prior samples during inference. The predicted class probability for query point $j$ is computed as the average over $T$ softmax-normalized outputs:
\begin{equation}
p_{j,c} = \frac{1}{T} \sum_{t=1}^T \mathrm{softmax}(\ell^{(t)}_{j,c}),
\label{eq:mc_avg}
\end{equation}
where $\ell^{(t)}_{j,c}$ denotes the similarity-based logit obtained using the $t$-th sampled prototype.

Uncertainty is estimated via two complementary measures. First, we compute prediction variance across samples, which captures epistemic uncertainty. Second, we optionally compute the entropy of the mean prediction distribution, reflecting overall confidence. Both metrics are fused to yield per-point uncertainty maps.

The overall training objective combines three components:
\begin{equation}
\mathcal{L} = \mathcal{L}_{\mathrm{seg}} + \mathcal{L}_{\mathrm{base}} + \beta(e)\, \mathcal{L}_{\mathrm{KL}},
\label{eq:loss}
\end{equation}
where $\mathcal{L}_{\mathrm{seg}}$ is the supervised segmentation loss on novel-class queries, $\mathcal{L}_{\mathrm{base}}$ is the auxiliary base-class supervision loss (following \cite{an2024rethinking}), and $\mathcal{L}_{\mathrm{KL}}$ is the KL divergence defined in Eq.~\ref{eq:kl}. The weight $\beta(e)$ follows a linear warm-up schedule with training epoch $e$.




\begin{figure}[tb]
  \centering
  \includegraphics[width=0.95\linewidth]{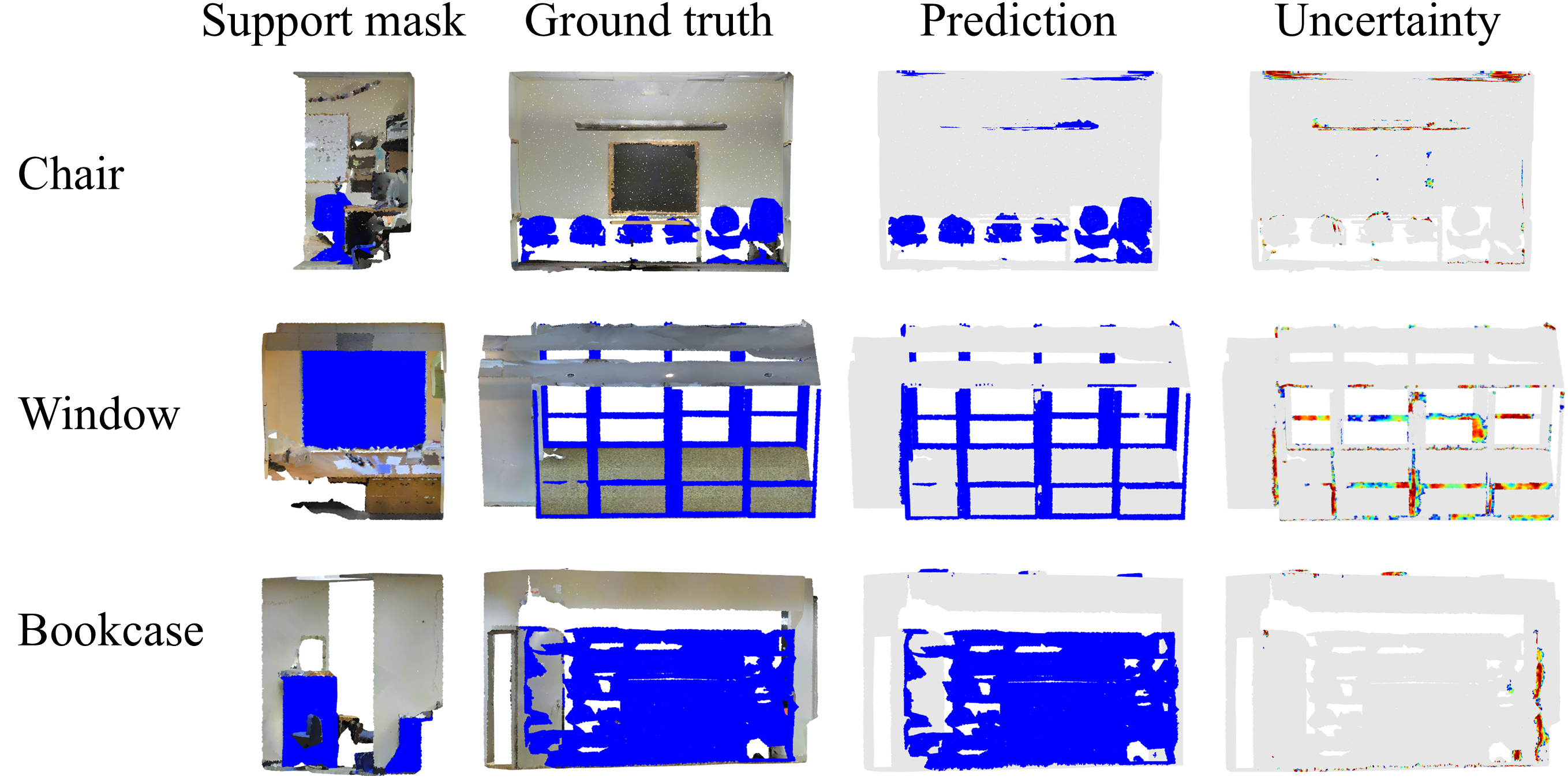}
  \vspace{-0.3cm} 
  \caption{Qualitative results of UPL on S3DIS. UPL yields not only accurate mask prediction but also reliable uncertainty estimation.}
  \vspace{-0.5cm} 
  \label{fig:vis}
\end{figure}
\section{Experiments}

\subsection{Experimental Setup}
\noindent\textbf{Datasets.} 
Following CoSeg~\cite{an2024rethinking}, we evaluate our model on two widely used FS-PCS benchmarks, i.e, S3DIS\cite{armeni20163d} and ScanNet\cite{dai2017scannet}. 
Both datasets are split into disjoint base and novel classes, and we adopt the episodic $N$-way $K$-shot setting. 
In each episode, $K$ support fragments and one query scene are sampled for $N$ novel classes.  

\noindent\textbf{Evaluation Metrics.} 
Following prior work~\cite{an2024rethinking}, we apply denser support sampling and improved class balancing to mitigate foreground leakage issues observed in earlier setups. 
We report mean Intersection over Union (mIoU, \%) over novel classes, averaged across two random seeds (S0/S1) and four few-shot settings: 1-way 1-shot, 1-way 5-shot, 2-way 1-shot, and 2-way 5-shot. 

\begin{figure*}[t]
  \centering
  \includegraphics[width=0.9\textwidth]{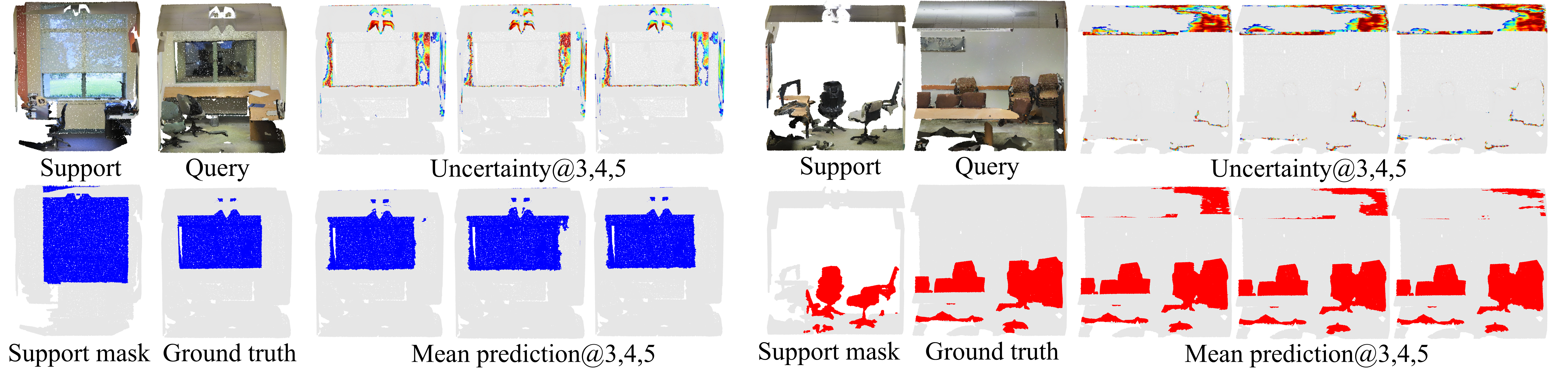}
  \vspace{-0.3cm} 
  \caption{Visualization of uncertainty analysis for the 1-way 1-shot setting. The left panel highlights the foreground class \textbf{window} (in blue), while the right panel focuses on the class \textbf{chair} (in red). Uncertainty is represented using a heatmap, where values transition from blue (low uncertainty) to red (high uncertainty). The number following @ denotes the sampling times $T$.}
  \vspace{-0.6cm} 
  \label{fig:uncertainty_vis}
\end{figure*}

\subsection{Main Results}
Table~\ref{tab:main_results} reports mIoU on S3DIS and ScanNet under the four standard few-shot settings (averaged over S0/S1). \method consistently outperforms all baselines on both datasets. 
On S3DIS, \method attains 48.60, 52.22, 37.79, and 41.87 mIoU, surpassing the strongest baseline (CoSeg) by up to +2.18 points. 
On ScanNet, it obtains 43.00, 46.83, 32.39, and 38.40 mIoU, with even larger gains in the 2-way settings (+3.61 and +3.72). 
These improvements are consistent across seeds and align with our design: (i) the dual-stream prototype refinement leverages query features to reduce intra-class variability and inter-class confusion, and the variational prototype inference regularization module that enables uncertainty-aware prototype learning.
Fig.~\ref{fig:vis} provides qualitative results on S3DIS. As shown in the figure, \method produces cleaner object boundaries and suppresses spurious activations in ambiguous regions (e.g., class boundaries and cluttered areas). 
More importantly, \method also outputs uncertainty maps alongside the segmentation masks, which clearly highlight regions of occlusion and label ambiguity. 
This not only improves the interpretability of the predictions but also demonstrates the reliability of our uncertainty-aware prototype learning framework.

\noindent\textbf{Effect of $K$.} 
As expected, increasing the number of support examples ($K$) consistently improves performance for all methods. 
Notably, \method benefits more from larger $K$. 
For example, on S3DIS the gain from 1w1s to 1w5s is +3.62 points, compared to only +2.83 for the strongest baseline. 
This indicates that \method adapts more effectively to intra-class variation, thanks to stochastic prototype sampling in the variational inference module.  

\noindent\textbf{Effect of $N$.} 
Increasing the number of novel classes ($N$) makes the task more challenging by introducing greater inter-class ambiguity. 
Although performance decreases for all methods, \method preserves a larger fraction of its 1-way accuracy than competing approaches. 
This demonstrates the effectiveness of dual-stream refinement and variational inference in mitigating class overlap and maintaining robust segmentation under more complex settings.

\begin{table}[tb]
\centering
\caption{Ablation study of DPR and VPIR modules on ScanNet (2-way). Performance is reported as mIoU (\%).}
\label{tab:ablation}
\small
\begin{tabular}{c|c|c|c}
\hline
\textbf{DPR} & \textbf{VPIR} & \textbf{2-way 1-shot} & \textbf{2-way 5-shot} \\
\hline
& & 27.32 & 33.65 \\
& \cmark & 30.19 & 36.04 \\
\cmark & \cmark & \textbf{32.39} & \textbf{38.40} \\
\hline
\end{tabular}
\vspace{-0.5cm} 
\end{table}
    
\subsection{Ablation Study}
\noindent\textbf{Benefits of DPR.}
Table~\ref{tab:ablation} reports the ablation study of the dual-stream refinement prototype attention (DPR) and variational prototype inference regularization(VPIR) modules on ScanNet under both 2-way 1-shot and 5-shot settings.
Both modules individually improve performance compared to the baseline without them.
Specifically, DPR alone brings notable gains (+5.07 and +4.75 mIoU in 1-shot and 5-shot, respectively), demonstrating the importance of jointly leveraging support and query features to construct more discriminative prototypes.

\noindent\textbf{Benefits of VPIR.}
VPIR alone also provides clear improvements (+2.87 and +2.39), validating the effectiveness of modeling prototypes as distributions for robust estimation.
When combined, DPR and VPIR achieve the best results (32.39 and 38.40), showing complementary benefits: refinement reduces intra/inter-class confusion, while variational inference further enhances robustness through uncertainty-aware prototype learning.

\noindent\textbf{Impact of Prior Sampling.}
Table~\ref{tab:T_mIoU} shows the effect of varying the number of prior samples $T$ during inference on the S3DIS dataset under 1-way settings.
As $T$ increases from 3 to 5, mIoU improves from 48.60\% (\textbf{Expected Calibration Error} 0.1710) to 51.69\% (0.0567) in the 1-shot case, while in the 5-shot case, the best performance (53.18\%) occurs at $T=4$.
These results demonstrate that leveraging multiple stochastic samples enhances prediction stability, though the optimal number of samples may vary with supervision strength.
The improvement reflects the ensemble-like benefits of Monte Carlo sampling, which mitigates prediction variance and leads to more reliable few-shot performance.

\noindent\textbf{Analysis of Uncertainty.}  
Fig.~\ref{fig:uncertainty_vis} visualizes the uncertainty maps produced by our framework. 
High uncertainty is primarily concentrated along object boundaries and in ambiguous regions, such as the \textbf{chair} and \textbf{window} classes. 
Importantly, incorrect predictions tend to correlate with higher uncertainty, which is clearly reflected by the shift from blue (low) to red (high) in the heatmaps. 
These observations highlight two key advantages of our probabilistic design: (i) it provides interpretable uncertainty estimates that can guide the identification of error-prone regions, and (ii) it improves prediction reliability, as increasing $T$ reduces uncertainty and corrects misclassifications.
Together, these results demonstrate that our method contributes both to performance gains and to model interpretability.

\begin{table}[tb]
\centering
\caption{Effect of the number of prior samples $T$ on S3DIS under 1-way settings. Performance is reported as mIoU (\%).}
\label{tab:T_mIoU}
\begin{tabular}{c|c|c}
\hline
\textbf{$T$} & \textbf{1-way 1-shot} & \textbf{1-way 5-shot} \\
\hline
3 & 48.60 & 52.22 \\
4 & 50.32 & 53.18 \\
5 & 51.69 & 53.06 \\
\hline
\end{tabular}
\vspace{-0.5cm} 
\end{table}

\section{Conclusion}
In this work, we introduced \textbf{UPL}, a probabilistic framework for few-shot 3D point cloud segmentation that enables uncertainty-aware prototype learning. 
UPL incorporates a dual-stream prototype refinement module to construct more discriminative prototypes, and a variational prototype inference regularization module to treat prototypes as latent variables, allowing explicit uncertainty modeling. 
Through extensive experiments on the S3DIS and ScanNet benchmarks, UPL achieves state-of-the-art performance while providing robust and interpretable uncertainty estimation. 
Our study highlights the importance of modeling uncertainty in few-shot point cloud segmentation. 
We believe this work opens up promising directions for extending uncertainty-aware learning to broader 3D tasks.

\newpage 
\section{Acknowledgments}
This research was supported by the National Key Research and Development Program of China (No.2023YFC3304800).
The computations in this research were performed using the CFFF platform of Fudan University.
Deployed on the Cross-Border Trade Payment Intelligent Agent Platform (\url{https://fdueblab.cn/}).

\bibliographystyle{IEEEbib}
\bibliography{refs_zyf}

\end{document}